\documentclass[letterpaper, 10 pt, conference]{ieeeconf}  

\IEEEoverridecommandlockouts                              

\usepackage{graphicx}
\usepackage{amsmath,amsfonts}
\usepackage[hidelinks]{hyperref}
\hypersetup
{
colorlinks = true,
linkcolor = red,
anchorcolor = black,
citecolor = green,
urlcolor = magenta
}

\title{\LARGE \bf
CMR-Agent: Learning a Cross-Modal Agent for Iterative \\ Image-to-Point Cloud Registration
}

\author{Gongxin Yao$^{1}$, Yixin Xuan$^{2}$, Xinyang Li$^{2}$ and Yu Pan$^{1,*}$
\thanks{$^{1}$Gongxin Yao and Yu Pan are with the Institute of Cyber-Systems and Control, Zhejiang University, Hangzhou, 310027, China. (Yu Pan* is the corresponding author, email: ypan@zju.edu.cn)}%
\thanks{$^{2}$Yixin Xuan and Xinyang Li are with the Polytechnic Institute, Zhejiang University, Hangzhou, 310015, China.}%
\thanks{This research is supported by the National Natural Science Foundation of
China under Grants No. U22A20102.}
}

\begin{document}

\maketitle
\thispagestyle{empty}
\pagestyle{empty}

\begin{abstract}

Image-to-point cloud registration aims to determine the relative camera pose of an RGB image with respect to a point cloud. It plays an important role in camera localization within pre-built LiDAR maps. Despite the modality gaps, most learning-based methods establish 2D-3D point correspondences in feature space without any feedback mechanism for iterative optimization, resulting in poor accuracy and interpretability. In this paper, we propose to reformulate the registration procedure as an iterative Markov decision process, allowing for incremental adjustments to the camera pose based on each intermediate state. To achieve this, we employ reinforcement learning to develop a cross-modal registration agent (CMR-Agent), and use imitation learning to initialize its registration policy for stability and quick-start of the training. According to the cross-modal observations, we propose a 2D-3D hybrid state representation that fully exploits the fine-grained features of RGB images while reducing the useless neutral states caused by the spatial truncation of camera frustum. Additionally, the overall framework is well-designed to efficiently reuse one-shot cross-modal embeddings, avoiding repetitive and time-consuming feature extraction. Extensive experiments on the KITTI-Odometry and NuScenes datasets demonstrate that CMR-Agent achieves competitive accuracy and efficiency in registration. Once the one-shot embeddings are completed, each iteration only takes a few milliseconds. [\href{https://github.com/y2w-oc/CMR-Agent/}{Code}].

\end{abstract}

\section{INTRODUCTION}

\begin{figure}[t]
    \centering
    \includegraphics[width=1.0\linewidth]{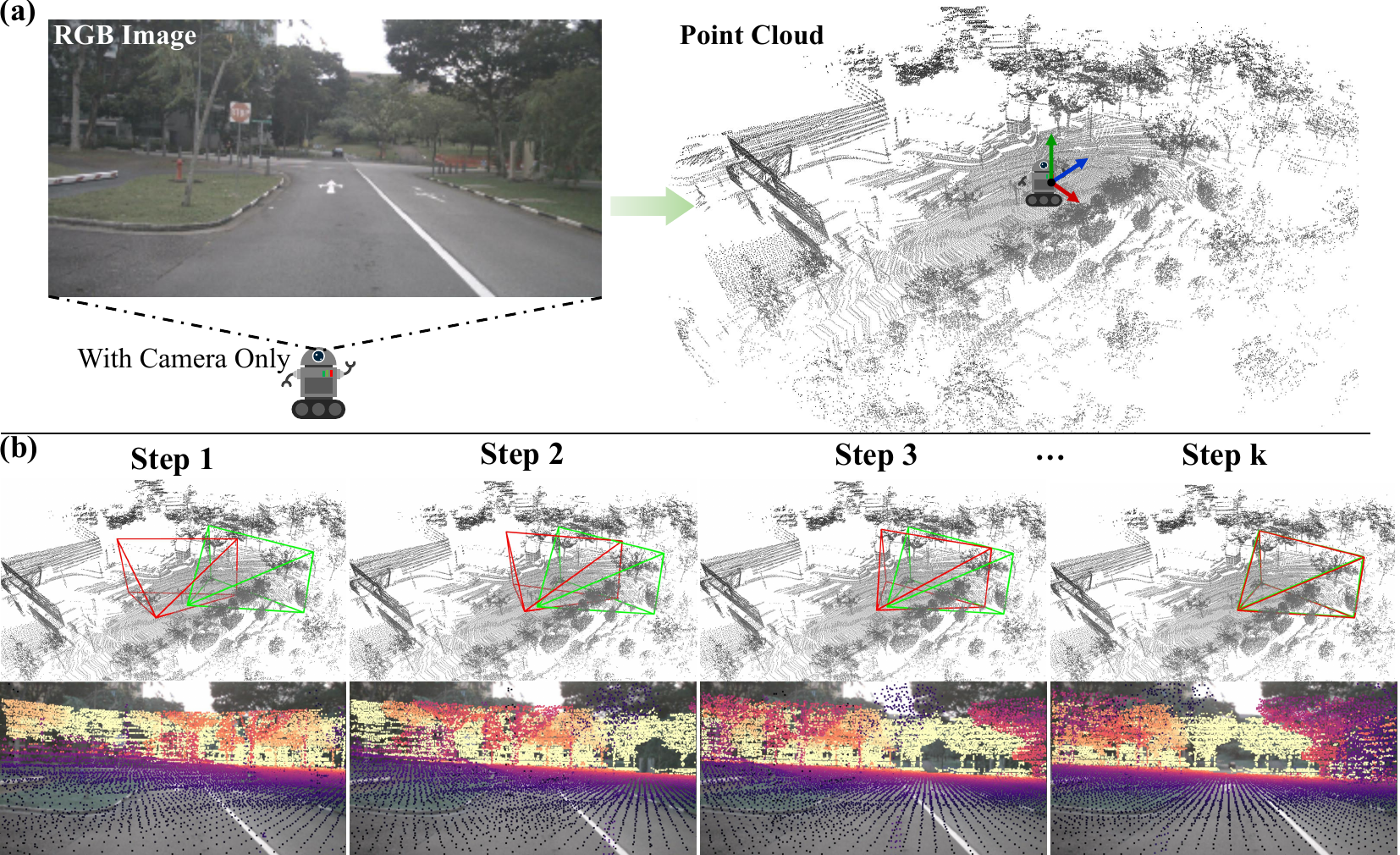}
    \vspace{-5mm}
    \caption{Illustrations of image-to-point cloud registration. (a) A lightweight system equipped solely with a camera can localize itself by registering the online RGB image and the pre-built LiDAR map. (b) Our CMR-Agent takes the cross-modal observations as inputs, predicting actions (i.e., relative rigid transformations) to iteratively improve registration. The red and green frustums represent the estimated and actual camera poses, respectively.}
    \label{fig1}
\end{figure}

Accurate localization mechanisms are crucial for autonomous driving systems. Although the Global Positioning System (GPS) \cite{hofmann2012global} is widely used, obstacles in real-world environments frequently disrupt the reception of satellite signals, affecting the precision and reliability of localization. To address these challenges, some methods eschew satellite communication and instead use sensors like LiDAR \cite{de2020evaluating} and camera \cite{chen2023deep} to localize by comparing online sensor measurements with reference maps. Specifically, LiDAR-based methods \cite{huang2021predator} estimate relative position and rotation by registering online 3D scans with a pre-built point cloud map. They have exhibited superior performance, primarily due to the accurate distance measurements and the immunity to variations in lighting and viewing angles. However, the substantial cost of LiDAR sensors restrict their deployment on micro platforms. Camera-based methods \cite{2021LoFTR} achieve localization by matching and aligning real-time RGB images with previously captured photographs. Despite the economic and lightweight advantages of cameras, these methods are notably susceptible to variations of environmental conditions, such as lighting, weather, and season. Therefore, a question naturally arises: Is it possible to combine the strengths of both LiDAR-based and camera-based techniques by establishing direct connections between RGB images and point cloud maps? As shown in Fig.\ref{fig1} (a), the autonomous driving system can be equipped solely with a low-cost camera for efficient localization in pre-built LiDAR maps, reducing the negative impact of environmental changes on visual-based techniques, since LiDAR maps remain unaffected by such visual discrepancies.

The core of cross-modal localization is to estimate the relative camera pose between an RGB image and a point cloud, commonly termed as image-to-point cloud registration \cite{li2021deepi2p}. Most current approaches \cite{feng20192d3d, ren2022corri2p, yao2023cfi2p, zheng2023i2prec} focus on establishing direct 2D-3D correspondences in feature space, where the feature matching process is conducted only once. These methods are categorized as one-shot registration due to their lack of a feedback mechanism for iterative optimization. Motivated by these concerns, we argue that mimicking the instinctive human process of aligning two objects may enhance the performance of image-to-point cloud registration. It involves iterative observation and adjustment, allowing us to continuously fine-tune the camera pose based on immediate feedback. Accordingly, we reformulate the registration problem as a Markov decision process with multi-step iterations. While reinforcement learning \cite{szepesvari2022algorithms} offers an appropriate model to depict this process, migrating it to the cross-modal context remains a complex task.

In this paper, we employ reinforcement learning to develop a \textbf{C}ross-\textbf{M}odal \textbf{R}egistration \textbf{Agent} (CMR-Agent) that operates as illustrated in Fig. \ref{fig1} (b). Our primary focuses are three-fold: 1) How to construct cross-modal state representations? 2) How to guide the training of CMR-Agent? 3) How to improve the efficiency in light of the high time complexity due to repeated iterations? For the first problem, we leverage the current camera pose to project point clouds into 2D grids for pixel-by-pixel feature comparison with RGB images. Since the camera's field of view is a truncated segment of the 3D space, projections of point clouds from certain poses may not contain any useful information for decision (these states within 2D space are termed as neutral states in Section \ref{sec:state}). To supplement more information, we predict the actual camera frustum in 3D space by point classification and compare it with the camera frustum of the current pose. The final state representation integrates the information from both 2D and 3D space. For the second problem, we use the ground-truth camera pose to back-project 2D pixels into 3D points and design a point-to-point alignment reward function. The total reinforcement loss follows the well-known proximal policy optimization (PPO) algorithm \cite{schulman2017ppo}, and imitation learning \cite{ly2020learning} is utilized to initialize the registration policy for quick-start of the training. For the last problem, we observe that a smaller relative pose does not lead to better feature embeddings for point clouds and images. Consequently, the overall framework of CMR-Agent is meticulously crafted to perform one-shot cross-modal embeddings, which are efficiently reused during iteration. It circumvents the repetitive and time-consuming feature extraction. In conclusion, the main contributions are:
\begin{itemize}
    \item We reformulate image-to-point cloud registration as a Markov Decision process and develop a fully cross-modal registration agent using reinforcement learning and imitation learning.
    \item We propose a 2D-3D hybrid state representation that efficiently exploits the fine-grained features of images while reducing the useless neutral states.
    \item We propose a point-to-point alignment reward between the 3D points and the back-projected 2D pixels to guide the training of the agent.
    \item We design an efficient framework for the agent to reuse the one-shot cross-modal embeddings, decreasing the time complexity of repeated iterations. 
\end{itemize}

Extensive experiments on the KITTI-Odometry \cite{geiger2013vision} and NuScenes \cite{caesar2020nuscenes} datasets show that CMR-Agent outperforms the state-of-the-art methods in registration accuracy. Despite being an iterative method, CMR-Agent demonstrates higher efficiency than some one-shot algorithms, executing 10 iterations in less than 68 milliseconds on a 3090 GPU.

\section{Related works}

\subsection{Registration between Same Modal Data}

Registration plays an important role in camera-based and LiDAR-based localization methods. Image-to-image registration tackles multi-view geometry by estimating the relative pose between two images. The absence of depth information makes the epipolar constraint \cite{zhou2021patch2pix} a classic family of methods for delineating the geometric relationship between pixels across images. Besides, Optical flow \cite{zhai2021optical, lu2023transflow} and direct feature matching \cite{2021LoFTR, xie2024deepmatcher} constitute another classic family of methods for visual pose estimation. Point cloud registration seeks to determine the rigid transformation between two point clouds, a process simplified by the presence of 3D geometric information. Methods like Iterative Closest Point (ICP) \cite{ICP1992} and its variants \cite{yang2013go} are the seminal works to solve this problem. In addition, learning-based feature matching \cite{huang2021predator, qin2023geotransformer} combined with SVD decomposition \cite{papadopoulo2000estimating} is also a standard paradigm for point cloud registration. While image-to-image and point cloud registration techniques have achieved satisfactory performance, LiDAR devices are not suitable for micro platforms, and cameras are not robust to changes in environmental conditions.

\subsection{Cross-Modal Registration}

Image-to-point cloud registration combines the advantages of both camera and LiDAR. One of the most popular pipelines is to get 2D-3D point correspondences and then estimate the camera pose using a RANSAC-based PnP solver. 2D3D-MatchNet \cite{feng20192d3d}, CorrI2P\cite{ren2022corri2p}, and CFI2P \cite{yao2023cfi2p} employ pseudo-siamese neural networks to establish the cross-modal correspondences from 3D points to 2D pixels. In addition, there are some matching-free algorithms. HyperMap\cite{HyperMap2021} converts point clouds into depth maps and proposes a late projection strategy to precompute and compress the 3D map features offline. DeepI2P\cite{li2021deepi2p} frames the registration as a two-stage process of point-wise classification and custom pose optimization. EFGHNet\cite{EFGHNet2022} utilizes a divide-and-conquer approach, estimating rotation and translation separately by aligning RGB images with 360$^{\circ}$ panoramic depth maps. However, most of these methods are one-shot techniques with limited accuracy. The design of an iterative paradigm has been seldom explored.

\subsection{Reinforcement Learning in Registration}

Reinforcement learning \cite{szepesvari2022algorithms} provides a framework to describe the iterative decision-making process. Recent visual pose estimation algorithms \cite{Krull_2017_CVPR, Shao_2020_CVPR} leverage reinforcement learning to develop agents that can register two RGB images. Specifically, these image-oriented agents employ a fully convolutional architecture for feature extraction and state representation, outputting actions to iteratively adjust the relative camera pose. As for reinforcement learning-powered point cloud registration, ReAgent \cite{bauer2021reagent} employs the PointNet architecture to represent states, progressively moving the source towards the target point clouds. In this study, our goal is to extend reinforcement learning to cross-modal registration between images and point clouds. Direct application of the previous single-modal state representations is not feasible for this task. Additionally, developing training strategies for the agent and improving efficiency pose greater challenges in cross-modal contexts.

\begin{figure*}[t]
    \centering
    \includegraphics[width=1\linewidth]{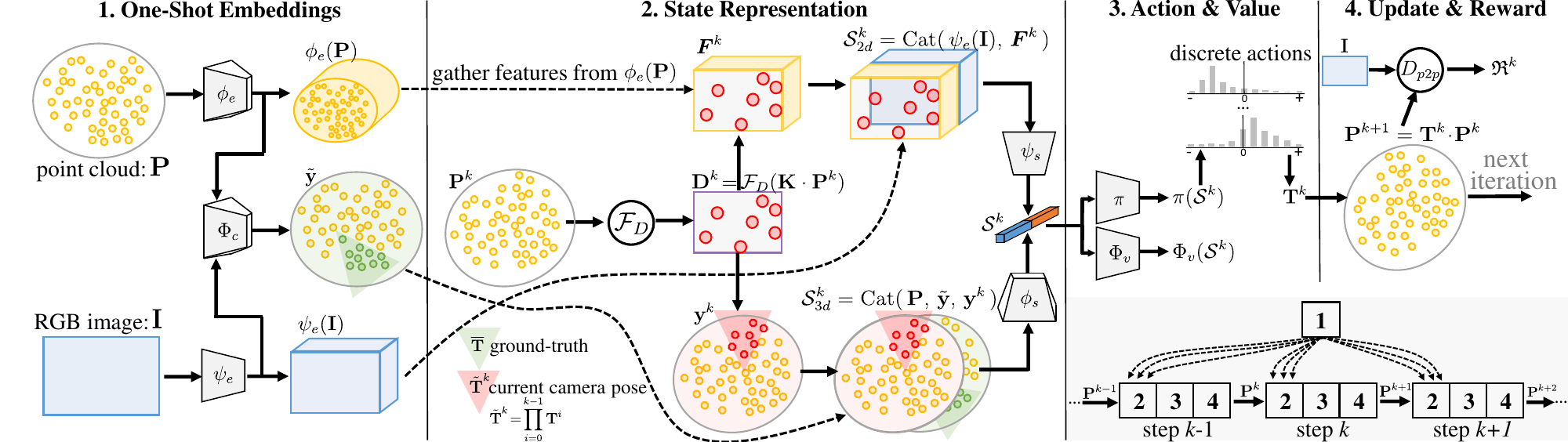}
    \vspace{-2mm}
    \caption{The overall framework of our cross-modal registration agent (CMR-Agent). Parts 2 to 4 on the right constitute the iterative body. The dashed lines indicate the data flow for reusing one-shot embeddings. The bottom right corner is a schematic overview of the entire iterative process.
    }
    \label{fig2}
\end{figure*}

\section{Methodology}

\subsection{Overview}
\label{sec:overview}

\textbf{Problem Formulation:} In the context of image-to-point cloud registration, denote the scene point cloud as $\textbf{P} \in \mathbb{R}^{N \times 3}$ with $N$ points and the RGB image as $ \textbf{I} \in \mathbb{R}^{H \times W \times 3} $ that is taken from the same scene, our goal is to estimate the camera pose $\overline{\textbf{T}} \in SE(3)$ when the image was taken. Formally, given the camera intrinsic matrix $\textbf{K} \in \mathbb{R}^{3\times3}$ and $\overline{\textbf{T}}$, we project $\textbf{P}$ to image plane and get a depth map $\overline{\textbf{D}} \in \mathbb{R}^{H \times W \times 1}$ as:
\begin{equation}
\begin{aligned}
    \label{eq1}
    [x_i\,, y_i\,, z_i]^{\mathsf{T}} &= \textbf{K} \cdot \overline{\textbf{T}} \cdot \textbf{p}_i \,\,,\\
    \textbf{u}_i &= [x_i / z_i\,, y_i / z_i]^{\mathsf{T}} ,\\
    \overline{\textbf{D}}_{\textbf{u}_i} &= z_i \,\,,
\end{aligned}
\end{equation}
where $\textbf{p}_i \in \textbf{P}$ is a 3D point and $\textbf{u}_i$ represents a 2D pixel. We encapsulate the entire process of deriving a depth map from the point cloud $\textbf{P}$ into a single function $\mathcal{F}_{D}(\cdot)$ as:
\begin{align}
    \label{eq2}
    \overline{\textbf{D}} &= \mathcal{F}_{D}(\textbf{K} \cdot \overline{\textbf{T}} \cdot \textbf{P}).
\end{align}
Although $\overline{\textbf{D}}$ and $\textbf{I}$ represent the scene using depth and color, respectively, they are aligned to each other pixel by pixel. Besides, we encapsulate the first two steps in Eq. (\ref{eq1}) into another function $\mathcal{F}_{u}(\cdot)$ for point-to-pixel projection as:
\begin{align}
    \label{eq3}
    \textbf{u}_i &= \mathcal{F}_{u}(\textbf{K} \cdot \overline{\textbf{T}} \cdot \textbf{p}_i).
\end{align}

\textbf{Markov Modeling:} Previous registration methods take $\textbf{P}$ and $\textbf{I}$ as the observations to derive a camera pose $\textbf{T} \in SE(3)$ in one step. Inevitably, $\textbf{T}$ shows great deviation from $\overline{\textbf{T}}$. In this study, we propose to reformulate the one-shot process as a Markov decision process, thereby progressively improving the estimated camera pose. To achieve this, we employ reinforcement learning to train a cross-modal registration agent $\mathcal{A}(\cdot)$. For every step $k \ge 0$, it takes $\textbf{P}^{k}$ and $\textbf{I}$ as the observations to predict a rigid transformation $\textbf{T}^{k} \! \in \! SE(3)$, and $\textbf{T}^{k}$ is then applied to $\textbf{P}^{k}$ to obtain $\textbf{P}^{k+1}$ as:
\begin{equation}
\begin{aligned}
    \label{eq4}
    \textbf{T}^{k} &= \mathcal{A}(\textbf{P}^{k}, \textbf{I}) ,\\
    \textbf{P}^{k+1} &= \textbf{T}^{k} \cdot \textbf{P}^{k} \,\,,
\end{aligned}
\end{equation}
where $\textbf{P}^{0} \! = \! \textbf{P}$. These two steps are executed alternately. The accumulation of rigid transformations before the $k$-th step is the current camera pose $\tilde{\textbf{T}}^k$, which we aim to progressively converge towards $\overline{\textbf{T}}$ as:
\begin{align}
    \label{eq5}
    \lim_{k \to \infty} \tilde{\textbf{T}}^k = \lim_{k \to \infty} \prod_{i=0}^{k-1} \textbf{T}^{i} = \overline{\textbf{T}}.
\end{align}
Note that each $\textbf{T}^k$ is termed as an action at the $k$-th step, and the calculation of $\tilde{\textbf{T}}^k$ follows the method of disentangled transformation \cite{li2018deepim} to avoid rotation-induced translation.

\textbf{Framework:} We implement $\mathcal{A}(\cdot)$ as a policy network, which uses state representation to predict the distribution of actions. $\mathcal{A}(\cdot)$ is more complex than the agents for same-modal registration \cite{Shao_2020_CVPR, bauer2021reagent} because $\textbf{P}^{k}$ and $\textbf{I}$ can't share neural network weights for processing. Although projecting $\textbf{P}^{k}$ as a depth map $\textbf{D}^{k} \! = \! \mathcal{F}_{D}(\textbf{K} \cdot \textbf{P}^{k})$ can unify the data structure \cite{cmrnet}, the disparate visual attributes between $\textbf{D}^{k}$ and $\textbf{I}$ require the network be deeper to achieve sufficient expressive capability, leading to high time complexity when iterating it. To this end, we propose an efficient framework for $\mathcal{A}(\cdot)$, as illustrated in Fig.\ref{fig1}. We first build a pseudo-siamese neural network to generate one-shot embeddings $\phi_e(\textbf{P}) \! \in \! \mathbb{R}^{N \times f}$ and $\psi_e(\textbf{I}) \! \in \! \mathbb{R}^{H \times W \times f}$ for $\textbf{P}$ and $\textbf{I}$, where $f$ is feature dimension. Instead of using $\textbf{P}^{k}$ and $\textbf{I}$ as the state representation $\mathcal{S}^k$ at the $k$-th step, we construct $\mathcal{S}^k$ by gathering features from $\phi_e(\textbf{P})$ and $\psi_e(\textbf{I})$ based on the current camera pose $\tilde{\textbf{T}}^k$. This design reuses $\phi_e(\textbf{P})$ and $\psi_e(\textbf{I})$ during iterations, avoiding repetitive and time-consuming feature extraction. After that, a policy head predicts the probability distribution of actions as $\pi(\mathcal{S}^k)$ for decision, and a value head predicts a value as $\Phi_v(\mathcal{S}^k)$ to guide the policy optimization. Once the current camera pose is updated, the agent receives a reward $\mathfrak{R}^k$ to judge how well it performs.


\subsection{One-Shot Cross-Modal Embeddings}
The goal of the networks $\phi_e(\cdot)$ and $\psi_e(\cdot)$ is to map $\textbf{P}$ and $\textbf{I}$ into a unified embedding space while making the correlated 2D pixels and 3D points closer. For this purpose, we adopt circle loss \cite{sun2020circle} to guide the training. We initially sample $n$ 3D points that fall in the camera frustum of pose $\overline{\textbf{T}}$. Denote the points as $\textbf{P}^{s} \in \mathbb{R}^{n \times 3}$ and $\textbf{P}^{s} \subset \textbf{P}$, the corresponding 2D pixels is $\textbf{U}^{s} = \{\textbf{u}_i \big| \textbf{u}_i =  \mathcal{F}_{u}(\textbf{K} \cdot \overline{\textbf{T}} \cdot \textbf{p}_i), \textbf{p}_{i} \in \textbf{P}^{s}\}$. For any 3D point $\textbf{p}_i$ , the set of positive 2D pixels is defined as $\mathcal{E}_{+}^{i} \! = \! \{\textbf{u}_{j} \,\big|\, \textbf{u}_{j} \in \textbf{U}^{s}, {\Vert \textbf{u}_{i} - \textbf{u}_{j}\Vert}_{2} \leq r  \}$, where $r$ is a radius. Conversely, $\mathcal{E}_{-}^{i} \! = \! \{\textbf{u}_{j} \,\big|\, \textbf{u}_{j} \in \textbf{U}^{s}, {\Vert \textbf{u}_{i} - \textbf{u}_{j}\Vert}_{2} > r \}$ is the negative set. Then, the circle loss anchored in 3D points is formulated as:
\begin{align}
    \label{eq6}
    \mathcal{L}^{3d}_{c} = \frac{1}{n} \sum_{i=1}^{n} \log \left[ 1 \! + \! \sum_{j\in\mathcal{E}_{+}^{i}} e^{\theta_+^j(d_i^j - \Delta_+)} \cdot \! \sum_{k\in\mathcal{E}_{-}^{i}} e^{\theta_-^k(\Delta_- - d_i^k)} \right],
\end{align}
where $d_i^j$ is the distance between $\textbf{p}_{i}$ and $\textbf{u}_{j}$ in the embedding space, and $\Delta_+$, $\Delta_-$ represent the positive and negative margins, respectively. The positive and negative weights are computed adaptively for each samples with $\theta_+^j = \lambda(d_i^j \! - \! \Delta_+)$ and $\theta_n^k = \lambda(\Delta_- \! - \! d_i^k)$. $\lambda$ is a hyper-parameter. Similarly, we get the reverse circle loss $\mathcal{L}^{2d}_{c}$ in the same way, and the total circle loss is defined as $\mathcal{L}_{c} \! = \! \mathcal{L}^{2d}_{c} \! + \! \mathcal{L}^{3d}_{c}$.

Additionally, we follow the previous work \cite{li2021deepi2p} to perform point-wise binary classification on $\textbf{P}$ as:
\begin{align}
    \label{eq7}
    \tilde{\textbf{y}} &= \Phi_c(\phi_e(\textbf{P}),\psi_e(\textbf{I})),
\end{align}
where $\Phi_c(\cdot)$ is the classification head, $\tilde{\textbf{y}} \in \mathbb{R}^{N \times 1}$ and any $\tilde{\textbf{y}}_i$ indicates whether the point $\textbf{p}_i$ is within the camera frustum of ground-truth pose $\overline{\textbf{T}}$. In 3D state representation, $\tilde{\textbf{y}}$ will be used to assess the frustum alignment. Note that the prediction of $\tilde{\textbf{y}}$ is also a one-shot step. Given $\overline{\textbf{T}}$ as a prior, we obtain the training labels via camera projection and checking whether a 3D point falls into the frustum. We adopt weighted binary cross-entropy as the loss to train $\Phi_c(\cdot)$.

\subsection{State Representation}
\label{sec:state}
At every step $k$, we first represent the state between $\textbf{P}^{k}$ and $\textbf{I}$ individually in image plane and point cloud space. Then, we merge the two into a hybrid state representation.

\textbf{2D State Representation:} We project the point cloud $\textbf{P}^{k}$ onto the image plane as $\textbf{D}^{k} \! = \! \mathcal{F}_{D}(\textbf{K} \cdot \textbf{P}^{k})$, and assess the alignment between $\textbf{D}^{k}$ and $\textbf{I}$. To improve the efficiency of assessment, we reuse the one-shot embeddings $\phi_e(\textbf{P})$ and $\psi_e(\textbf{I})$ here. Specifically, we first gather the 3D features from $\phi_e(\textbf{P})$ to fill the grid of $\textbf{D}^{k}$ and obtain $\boldsymbol{F}^k \! \in \! \mathbb{R}^{H \times W \times f}$. For every pixel $\textbf{u}$, we calculate $\boldsymbol{F}^k_{\textbf{u}} \in \mathbb{R}^f$ in $\boldsymbol{F}^k$ as:
\begin{equation}
\begin{aligned}
    \label{eq8}
    G^{k}_{\textbf{u}} &= \left\{ \textbf{p}_{i} \in \textbf{P}^{k} \,\big|\, \textbf{u} = \mathcal{F}_{u}(\textbf{K} \cdot \textbf{p}_i)\right\}, \\
    \boldsymbol{F}^k_{\textbf{u}} &= \sum\nolimits_{\textbf{p}_{i} \in G^{k}_{\textbf{u}}} \frac{\phi_e(\textbf{p}_i)}{|G^{k}_{\textbf{u}}|},
\end{aligned}
\end{equation}
where $\phi_e(\textbf{p}_i) \in \mathbb{R}^f$ is the feature of point $\textbf{p}_i$ from $\phi_e(\textbf{P})$ and $|\cdot|$ represents set cardinality. If $|G^{k}_{\textbf{u}}| = 0$, we initialize $\boldsymbol{F}^k_{\textbf{u}}$ as a zero vector. Then, the 2D state $\mathcal{S}^k_{2d} \! \in \! \mathbb{R}^{H \times W \times 2f}$ is represented as:
\begin{align}
    \label{eq9}
    \mathcal{S}^k_{2d} = \mathrm{Cat}(\, \psi_e(\textbf{I}),\, \boldsymbol{F}^k\,),
\end{align}
where $\mathrm{Cat}(\cdot)$ refers to the concatenate operation.

\begin{figure}[t]
    \centering
    \includegraphics[width=1.0\linewidth]{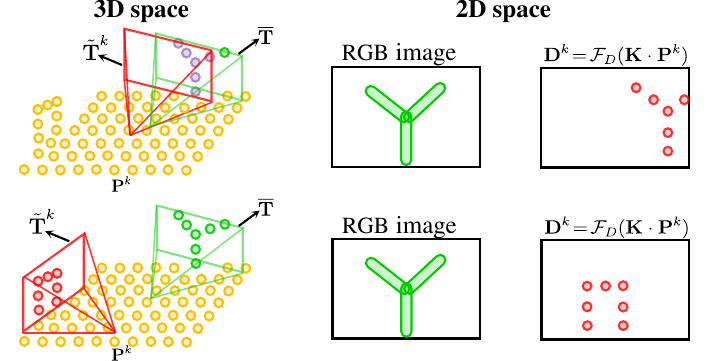}
    \vspace{-7mm}
    \caption{Illustrations of the useful states (top) and neutral states (bottom). The top row shows a fine case where the camera frustums of $\overline{\textbf{T}}$ and $\tilde{\textbf{T}}^k$ overlap with each other. After projection, the geometric features in the depth map $\textbf{D}^{k}$ can indicate actions to adjust the camera pose to align with the RGB image. In the bottom row, $\textbf{D}^{k}$ does not contain any geometric information of the RGB image, which is useless for action prediction.}
    \label{fig3}
\end{figure}

\textbf{3D State Representation:} We found that the 2D state representation is flawed. Because the camera frustum is a truncated segment of the 3D space, there are always certain poses that result in bad cases, i.e., no 3D point associated with the scene in $\textbf{I}$ is contained in the current camera frustum, as illustrated in Fig. \ref{fig3}. In such cases, the 2D states fail to provide any useful information for predicting the action policy, so we term these states as \textbf{neutral states}. To supplement more information, we assess the alignment between the camera frustums of $\tilde{\textbf{T}}^k$ and $\overline{\textbf{T}}$ in 3D space, as illustrated in Fig. \ref{fig2}. The frustum of camera pose $\overline{\textbf{T}}$ is predicted in Eq. (\ref{eq7}). Similarly, we also represent the current camera frustum of $\tilde{\textbf{T}}^k$ as binary classification results $\textbf{y}^k \! \in \! \mathbb{R}^{N \times 1}$, where every $\textbf{y}^k_i$ can be determined by camera projection and checking whether $\textbf{p}_i \! \in \! \textbf{P}^k$ falls into the current frustum. Then, the 3D state $\mathcal{S}^k_{3d} \! \in \! \mathbb{R}^{N \times 5}$ is represented as:
\begin{align}
    \label{eq10}
    \mathcal{S}^k_{3d} = \mathrm{Cat}(\, \textbf{P},\, \tilde{\textbf{y}}, \,\textbf{y}^k\,),
\end{align}

\textbf{2D-3D Hybrid State Representation:} To fully exploit the fine-grained features of $\mathcal{S}^k_{2d}$ while reducing the neutral states, we propose to merge both $\mathcal{S}^k_{2d}$ and $\mathcal{S}^k_{3d}$ into a hybrid state representation $\mathcal{S}^k \! \in \! \mathbb{R}^{2f'}$ as:
\begin{align}
    \label{eq11}
    \mathcal{S}^k = \mathrm{Cat}(\, \psi_{s}(\mathcal{S}^k_{2d}),\, \phi_{s}(\mathcal{S}^k_{3d})\,),
\end{align}
where $\psi_{s}(\mathcal{S}^k_{2d}) \! \in \! \mathbb{R}^{f'}$ and $\phi_{s}(\mathcal{S}^k_{3d}) \! \in \! \mathbb{R}^{f'}$. $\phi_{s}(\cdot)$ and $\psi_{s}(\cdot)$ are two networks that map the original 2D and 3D states to $d$ dimensional state space.

\subsection{Action Policy}
\label{sec:action}
The continuous action space couples $n_r \! + \! n_t$ degrees of freedom, e.g.., $n_r\!=\!3$ for rotation and $n_t\!=\!3$ for translation. Following the previous method \cite{Shao_2020_CVPR}, we decompose the action space into $n_r \! + \! n_t$ orthogonal subspace, and predict the action in each subspace with discrete and limited step size. Taking the $i$-th subspace as an example, our agent predicts the optimal action from a set of candidate actions $\textbf{A}_i \! = \! [-\Delta_{j}, -\Delta_{j-1}, \dots, 0, \dots, +\Delta_{j-1}, +\Delta_{j}]$, where $N_a \! = \! |\textbf{A}_i|$ (i.e., $N_a $ is $2j+1$ here). Note that the rotation actions are expressed by the angles around different axes. The signs $+$ and $-$ indicate the direction of the action. Thus, given the state $\mathcal{S}^k$, our agent will predict the action policy as $\pi(\mathcal{S}^k) \! \in \! \mathbb{R}^{(n_r \! + \! n_t) \times N_a}$, which are the probability distributions of the candidate actions in all subspace. In the $i$-th subspace, the final action $\textbf{a}_i^k$ is calculated as:
\begin{align}
    \label{eq12}
    \textbf{a}_i^k = \mathop{\mathrm{arg\,max}}\limits_{\textbf{a} \in \textbf{A}_i} \, \pi(\textbf{a} | \mathcal{S}^k),
\end{align}
where $\pi(\textbf{a} | \mathcal{S}^k)$ represents the probability of action $\textbf{a}$ from $\pi(\mathcal{S}^k)$. Finally, we adopt the disentangled transformation method \cite{li2018deepim} to convert the $n_r \! + \! n_t$ actions into $\textbf{T}^{k}$, thereby updating the current camera pose.

\subsection{Policy Optimization}

\begin{figure}[t]
    \centering
    \includegraphics[width=1.0\linewidth]{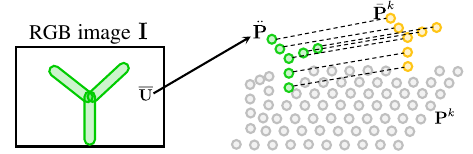}
    \vspace{-6mm}
    \caption{Illustration of the point-to-point distance between the RGB image and the point cloud.}
    \label{fig4}
\end{figure}

\textbf{Reward:} The goal of the cross-modal registration agent is to align the image and the point cloud as described in Section \ref{sec:overview}, therefore, the degree of alignment can serve as a criterion for evaluating the action quality. To quantify it, an intuitive approach is to calculate the average distance between the corresponding pixels in $\textbf{D}^{k} \! = \! \mathcal{F}_{D}(\textbf{K} \cdot \textbf{P}^{k})$ and $\textbf{I}$, however, it does not work in the neutral states. Instead, we calculate an average distance in 3D space, as illustrated in Fig. \ref{fig4}. Denote the 3D points that fall in the camera frustum of pose $\overline{\textbf{T}}$ as $\bar{\textbf{P}} \in \mathbb{R}^{m \times 3}$ and $\bar{\textbf{P}} \subset \textbf{P}$, they are transformed as $\bar{\textbf{P}}^{k} \! = \! \tilde{\textbf{T}}^k \cdot \bar{\textbf{P}}\! $ \,at step $k$. The corresponding 2D pixels are defined as $\overline{\textbf{U}} = \{\textbf{u}_i \big| \textbf{u}_i =  \mathcal{F}_{u}(\textbf{K} \cdot \overline{\textbf{T}} \cdot \textbf{p}_i), \textbf{p}_{i} \in \bar{\textbf{P}}\}$. We back-project $\overline{\textbf{U}}$ to 3D space as $\Ddot{\textbf{P}}$ through the reverse process of Eq. (\ref{eq1}). Then, the average distance between $\Ddot{\textbf{P}}$ and $\bar{\textbf{P}}^{k}$ is:
\begin{align}
    \label{eq13}
    D_{p2p}(\Ddot{\textbf{P}}, \bar{\textbf{P}}^k) = \sum\nolimits_{i=1}^{m} \frac{{\Vert \Ddot{\textbf{P}}_i - \bar{\textbf{p}}^k_i \Vert}_{2}}{m}\,.
\end{align}
During the process of iterative registration, this distance will become smaller until 0. Inspired by the previous work \cite{bauer2021reagent}, the step-wise reward is defined as:
\begin{equation}
    \label{eq14}
    \mathfrak{R}^k = \begin{cases}
        \varepsilon^- & \mathrm{if}  \ D_{p2p}(\Ddot{\textbf{P}}, \bar{\textbf{P}}^{k+1}) > D_{p2p}(\Ddot{\textbf{P}}, \bar{\textbf{P}}^{k})\,, \\
	\varepsilon^o & \mathrm{if} \ D_{p2p}(\Ddot{\textbf{P}}, \bar{\textbf{P}}^{k+1}) = D_{p2p}(\Ddot{\textbf{P}}, \bar{\textbf{P}}^{k})\,,  \\
        \varepsilon^+ & \mathrm{if} \ D_{p2p}(\Ddot{\textbf{P}}, \bar{\textbf{P}}^{k+1}) < D_{p2p}(\Ddot{\textbf{P}}, \bar{\textbf{P}}^{k})\,,
    \end{cases}
\end{equation}
where $\varepsilon^- \! \le \! 0$ and $\varepsilon^o \! \le \! 0$ penalize pausing and divergent actions, and $\varepsilon^+ \! > \! 0$ encourages actions that promote alignment.

\textbf{Optimization:} Our agent is based on actor-critic architecture \cite{szepesvari2022algorithms} and we adopt the classic proximal policy optimization (PPO) algorithm \cite{schulman2017ppo} to optimize the policy. For this purpose, there is also a value head to predict $\Phi_v(\mathcal{S}^k)$, and the Generalized Advantage Estimation (GAE) \cite{schulman2015high} is adopted to calculate advantage based on the reward and the value. Following common practices \cite{szepesvari2022algorithms}, we rollout the trajectories of iterative registration by executing the stochastic policy of the agent and record them in a replay buffer. During training, we gather a batch of random samples from the buffer to update the parameters of agent. The total PPO loss consists of three components: a policy loss for optimizing the policy head $\pi(\cdot)$, a value loss for optimizing the value head $\Phi_v(\cdot)$, and an entropy loss of the policy to encourage exploration. For readers unfamiliar with PPO, we recommend referring to the original paper \cite{schulman2017ppo} for details.

\subsection{Quick-Start with Imitation Learning}
Learning a cross-modal registration agent from scratch with reinforcement learning will take a long time to converge and might settle into a suboptimal policy. To address these problems, we employ imitation learning \cite{ly2020learning} to initialize the action policy. As the name suggests, the goal of imitation learning is to mimic the behavior of expert, where the expert could be either a human or another algorithmic approach. Here, we adopt the simplest form of imitation learning, Behavioral Cloning, which takes expert actions as training labels to supervise the action policy of agent. Given $\overline{\textbf{T}}$ as a prior during training, we can easily obtain a greedy expert policy. Specifically, we first calculate the difference between $\overline{\textbf{T}}$ and the current camera pose $\tilde{\textbf{T}}^k$ on the fly as:
\begin{align}
    \label{eq15}
    \Delta_{\textbf{T}} &= \overline{\textbf{T}} \cdot (\tilde{\textbf{T}}^k)^{-1}.
\end{align}
After that, we decompose $\Delta_{\textbf{T}}$ into the optimal actions in $n_r \! + \! n_t$ subspace as described in Section \ref{sec:action}. In the $i$-th subspace, our greedy expert policy selects the action from the discrete set $\textbf{A}_i$ that is closest to the optimal action mentioned above. Apparently, this is a multi-classification problem, so the imitation learning loss is the cross-entropy between the agent policy $\pi(\mathcal{S}^k)$ and the greedy expert policy.

\begin{table*}[t]
\centering
\caption{Quantitative comparison of registration performance. The best results in each column are highlighted in bold.}
\vspace{-2mm}
\label{tab1}
\renewcommand\arraystretch{1.2}
\begin{tabular}{c|ccc|ccc|ccc}
\hline
                & \multicolumn{3}{c|}{KITTI-Odometry-09 (3182 Samples)}                              & \multicolumn{3}{c}{KITTI-Odometry-10 (2402 Samples)}                               & \multicolumn{3}{|c}{NuScenes (16225 Samples)}\\ \hline
                & \multicolumn{1}{c|}{RTE(m)}      & \multicolumn{1}{c|}{RRE($^\circ$)}      & RR(\%) & \multicolumn{1}{c|}{RTE(m)}       & \multicolumn{1}{c|}{RRE($^\circ$)}      & RR(\%) & \multicolumn{1}{c|}{RTE(m)}       & \multicolumn{1}{c|}{RRE($^\circ$)}      & RR(\%)\\ \hline
GridCls \cite{li2021deepi2p}         & \multicolumn{1}{c|}{1.199$\pm$0.724} & \multicolumn{1}{c|}{6.000$\pm$1.989} & 76.39  & \multicolumn{1}{c|}{1.063$\pm$0.566}  & \multicolumn{1}{c|}{5.744$\pm$2.033} & 85.22  & \multicolumn{1}{c|}{2.224$\pm$1.094}  & \multicolumn{1}{c|}{7.370$\pm$1.647} & 62.67\\ 
DeepI2P(3D) \cite{li2021deepi2p}     & \multicolumn{1}{c|}{1.226$\pm$0.757} & \multicolumn{1}{c|}{6.113$\pm$2.279} & 39.69  & \multicolumn{1}{c|}{1.178$\pm$0.735}  & \multicolumn{1}{c|}{6.040$\pm$2.316} & 36.55  & \multicolumn{1}{c|}{1.730$\pm$0.969}  & \multicolumn{1}{c|}{7.058$\pm$2.184} & 18.82\\ 
DeepI2P(2D) \cite{li2021deepi2p}     & \multicolumn{1}{c|}{1.435$\pm$0.918} & \multicolumn{1}{c|}{3.805$\pm$2.660} & 79.73  & \multicolumn{1}{c|}{1.389$\pm$0.862}  & \multicolumn{1}{c|}{3.990$\pm$2.727} & 67.57  & \multicolumn{1}{c|}{1.949$\pm$1.050}  & \multicolumn{1}{c|}{3.017$\pm$2.285} & 92.53\\ 
EFGHNet \cite{EFGHNet2022}         & \multicolumn{1}{c|}{3.040$\pm$1.269} & \multicolumn{1}{c|}{4.526$\pm$2.347} & 24.18  & \multicolumn{1}{c|}{3.062$\pm$1.297} & \multicolumn{1}{c|}{4.706$\pm$2.371} & 23.57  & \multicolumn{1}{c|}{3.918$\pm$1.493}  & \multicolumn{1}{c|}{5.742$\pm$3.439} & 31.25\\ 
CorrI2P \cite{ren2022corri2p}        & \multicolumn{1}{c|}{0.919$\pm$0.742} & \multicolumn{1}{c|}{2.417$\pm$1.988} & 92.65  & \multicolumn{1}{c|}{0.878$\pm$0.689}  & \multicolumn{1}{c|}{3.126$\pm$2.333} & 91.59  & \multicolumn{1}{c|}{1.697$\pm$1.030}  & \multicolumn{1}{c|}{2.318$\pm$1.868} & 93.87\\ 
CFI2P \cite{yao2023cfi2p}          & \multicolumn{1}{c|}{0.610$\pm$0.471} & \multicolumn{1}{c|}{1.361$\pm$1.076} & 99.18  & \multicolumn{1}{c|}{0.574$\pm$0.483}  & \multicolumn{1}{c|}{1.438$\pm$1.076} & \textbf{99.63}  & \multicolumn{1}{c|}{1.093$\pm$0.668}  & \multicolumn{1}{c|}{1.465$\pm$1.230} & \textbf{99.23}\\  \hline
CMR-Agent & \multicolumn{1}{c|}{\textbf{0.195$\pm$0.151}} & \multicolumn{1}{c|}{\textbf{0.589$\pm$0.843}} & \textbf{99.94}  & \multicolumn{1}{c|}{\textbf{0.247$\pm$0.210}}  & \multicolumn{1}{c|}{\textbf{0.738$\pm$1.089}} & \textbf{99.63}  & \multicolumn{1}{c|}{\textbf{1.007$\pm$0.680}}  & \multicolumn{1}{c|}{\textbf{0.982$\pm$1.337}} & 96.96\\ \hline
\end{tabular}
\end{table*}

\begin{figure}[t]
    \centering
    \includegraphics[width=1\linewidth]{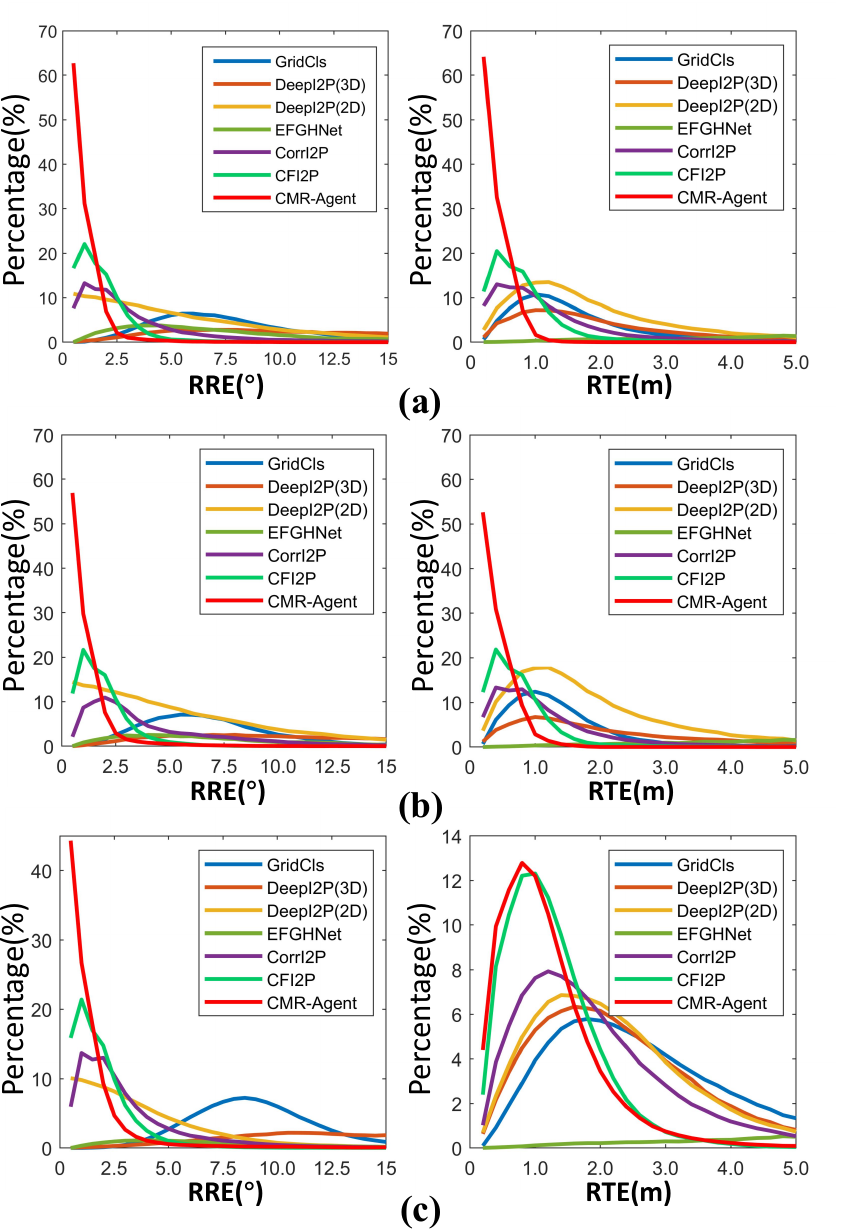}
    \vspace{-7mm}
    \caption{Error distributions on two datasets. (a) The sequence 09 of KITTI-Odometry. (b) The sequence 10 of KITTI-Odometry. (c) NuScenes.
    }
    \label{fig5}
\end{figure}


\begin{table}[t]
\centering
\caption{Computational Resources: Parameters, Running Memory, and Inference Time. The best results are highlighted in bold.}
\vspace{-2mm}
\label{tab2}
\begin{tabular}{c|c|c|c}
\hline
                & Param.(MB)      & Mem.(GB)      & Time(s) \\ \hline
GridCls \cite{li2021deepi2p}        & 100.75 & 2.41 & \textbf{0.049}  \\ 
DeepI2P(3D) \cite{li2021deepi2p}     & 100.12 & 2.01 & 16.584  \\ 
DeepI2P(2D) \cite{li2021deepi2p}     & 100.12 & 2.01 & 9.381  \\ 
EFGHNet \cite{EFGHNet2022}        & 574.32 & 2.76 & 0.319  \\ 
CorrI2P \cite{ren2022corri2p}        & 141.07 & 2.88 & 0.088  \\ 
CFI2P \cite{yao2023cfi2p}          & 34.74 & 2.18 & 0.084  \\ \hline
CMR-Agent & \textbf{34.65} & \textbf{1.87} & 0.068  \\ \hline 
\end{tabular}
\end{table}

\section{Experiments}
\subsection{Experimental Settings}
\textbf{Implementation Details:} All the experiments are conducted on an NVIDIA RTX 3090 GPU. We implement our cross-modal agent with PyTorch 1.12, and train it using the Adam optimizer with an initial learning rate of 0.001 and a momentum of 0.98. The dimension of embedding space $f$ is 64. In the one-shot embedding, the positive sample radius $r$ is 1 pixel, the number of sampled points $n$ is 512, $\lambda$ is 10, $\Delta_+$ is 0.1, and $\Delta_-$ is 1.4. The dimension of hybrid state space $f'$ is 128. In action policy, we suggest using actions with exponential step sizes to quickly cover a large space during early iterations while allowing fine-tuning during later iterations. The set of candidate rotation actions is $0.1^\circ\times[0, \pm1, \pm5, \pm25, \pm125, \pm625]$, the set of candidate translation actions is $0.1m\times[0, \pm1, \pm3, \pm9, \pm27, \pm81]$, and the number of candidates in each subspace $N_a$ is 11. In the reward function, we set $(\varepsilon^-, \varepsilon^o, \varepsilon^+)$ to (-0.5, 0, 0.5) empirically. As for training, since CMR-Agent is agnostic to the one-shot cross-modal embedding, we choose to pre-train the backbone separately. Besides, we follow the work \cite{bauer2021reagent} to jointly optimize the PPO and imitation learning loss. 


\textbf{Dataset:} We conducted experiments on KITTI-Odometry \cite{geiger2013vision} and NuScenes \cite{caesar2020nuscenes} datasets. For KITTI-Odometry, sequences 0-8 are used for training, with sequences 9-10 used for testing. Each data pair consists of a single LiDAR scan as the input point cloud and an RGB image sharing the same frame ID. Each point cloud is downsampled to 40,960 points, and the image resolution is cropped to $160\times512\times3$. For the NuScenes dataset, its official split assigns 850 scenes for training and 150 scenes for testing. Here, multiple LiDAR scans proximate to the frame ID are merged to construct a more comprehensive and uniform point cloud map, alongside selecting the corresponding RGB image. Similarly, this point cloud is sampled to 40,960 points, but the image resolution is cropped to $160\times320\times3$. For a fair comparison, we follow the previous settings \cite{ren2022corri2p} to apply a random transformation to the point cloud, consisting of a rotation around the up-axis of up to 360$^{\circ}$ and a translation on the ground within a 10-meter range. It means that $n_r$ and $ n_t$ are 1 and 2, respectively. Note that the degrees of freedom can be easily extended to 6D by adjusting the action prediction head.

\textbf{Evaluation Metrics:} We follow the previous work \cite{ren2022corri2p} to quantify the performance of image-to-point cloud registration with three metrics: 1) Relative Rotation Error (RRE), 2) Relative Translation Error (RTE), and 3) Registration Recall (RR). RR is the proportion of successful registrations within the test set, with a registration deemed successful when the RRE is less than $\tau_{r}$ and the RTE is less than $\tau_{t}$.

\subsection{Comparison with the State-of-the-Arts}
\textbf{Baselines:} We compare our CMR-Agent with 6 recent algorithms, including the three variants of DeepI2P \cite{li2021deepi2p}, EFGHNet \cite{EFGHNet2022}, CorrI2P \cite{ren2022corri2p} and CFI2P \cite{yao2023cfi2p}. Note that EFGHNet is a method based on 360$^{\circ}$ panoramic depth maps, which require good initialization, yet our setup (i.e., 10-meter translation) surpasses the range within which it can work normally. Thus, its performance significantly diverges from what was reported in the original paper \cite{EFGHNet2022}.

\begin{figure*}[t]
    \centering
    \includegraphics[width=1\linewidth]{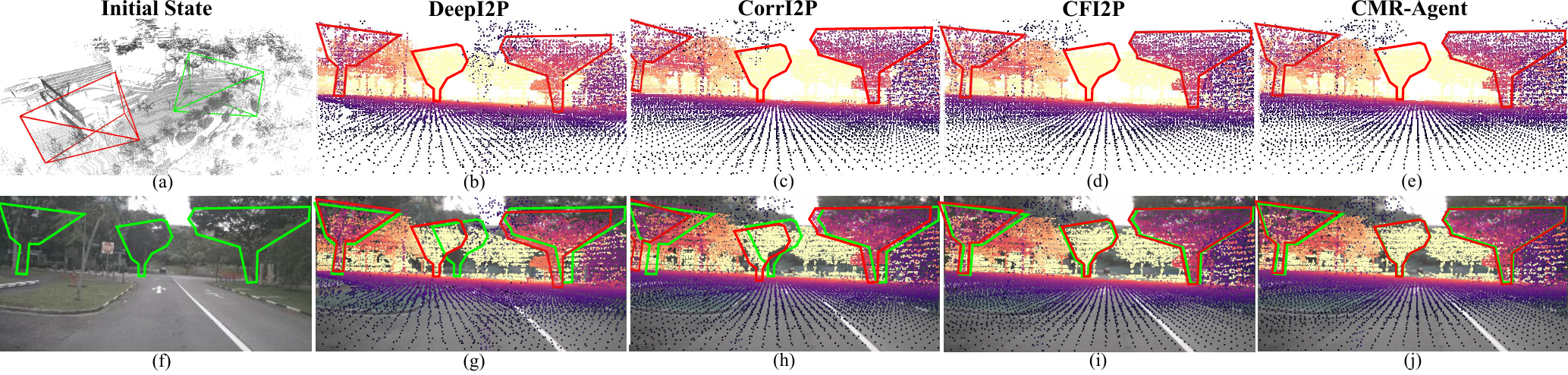}
    \vspace{-7mm}
    \caption{Qualitative results. (a) The initial (red) and ground-truth (green) camera poses are displayed in the point cloud. (b-e) The point projection maps based on the camera poses estimated by different algorithms. (f) The input RGB image. (g-j) Overlay results of the point projection maps and the RGB image. To clearly compare the alignment, we highlight some objects under the ground-truth (green) and the estimated (red) camera poses.
    }
    \label{fig6}
\end{figure*}

\textbf{Quantitative Comparison:} We report the registration performance of all methods in Table \ref{tab1}, where the results of CMR-Agent listed here are from its 10-th iteration, and the thresholds $\tau_{r}$ and $\tau_{t}$ of RR are set to $10^{\circ}$ and $5$ m, respectively. These statistical results clearly indicate that our CMR-Agent outperforms the state-of-the-art methods by a large margin. On the KITTI-Odometry dataset, CMR-Agent achieves the highest RR, with an average reduction of about 63\% in RRE and about 53\% in RTE compared to the suboptimal method CFI2P. On the NuScenes dataset, although RR is slightly lower than that of CFI2P, RRE is reduced by about 8\% and RTE is reduced by about 33\% on average. For a more comprehensive comparison, we plot the error distributions in Fig. \ref{fig5}, with $\tau_{r}$ ranging from 0 to $15^{\circ}$ and $\tau_{t}$ from 0 to 5 m. The results indicate that our method maintains higher accuracy even with smaller thresholds.

\textbf{Qualitative Comparison:} Image-to-point cloud registration is equivalent to the process of aligning 2D pixels with their corresponding 3D points from the same object. For a more intuitive comparison, we provide a visual example in Fig. \ref{fig6}. Owing to our CMR-Agent's ability to estimate relative camera poses with minimal error, it also outperforms other methods in achieving superior visual alignment.

\textbf{Efficiency:} The computational resources needed for all methods when executed on the KITTI-Odometry dataset are detailed in Table \ref{tab2}. Our CMR-Agent has the minimal number of parameters and the lowest memory footprint during execution. Remarkably, as an iterative method, CMR-Agent completes 10 iterations in just 68 milliseconds, making it the second fastest even when compared to the one-shot methods.


\begin{table}[t]
\centering
\caption{Contribution of individual components. The results are averaged over two sequences of KITTI-Odometry.}
\vspace{-2mm}
\label{tab3}
\begin{tabular}{c|c|c|c}
\hline
Method     & RTE(m)      & RRE($^\circ$)      & RR(\%) \\ \hline
w/o 3D State     & 0.292$\pm$0.294 & 0.882$\pm$1.176 & 97.64   \\ 
w/o 2D State     & 0.526$\pm$0.352 & 2.107$\pm$1.733 & 99.27   \\ 
w/o RL Supervision & 0.260$\pm$0.216 & 0.891$\pm$1.140 & 99.64 \\ 
Full Pipeline    & 0.217$\pm$0.181 & 0.653$\pm$0.959 & 99.80   \\ \hline
\end{tabular}
\end{table}

\begin{figure}[t]
    \centering
    \includegraphics[width=1\linewidth]{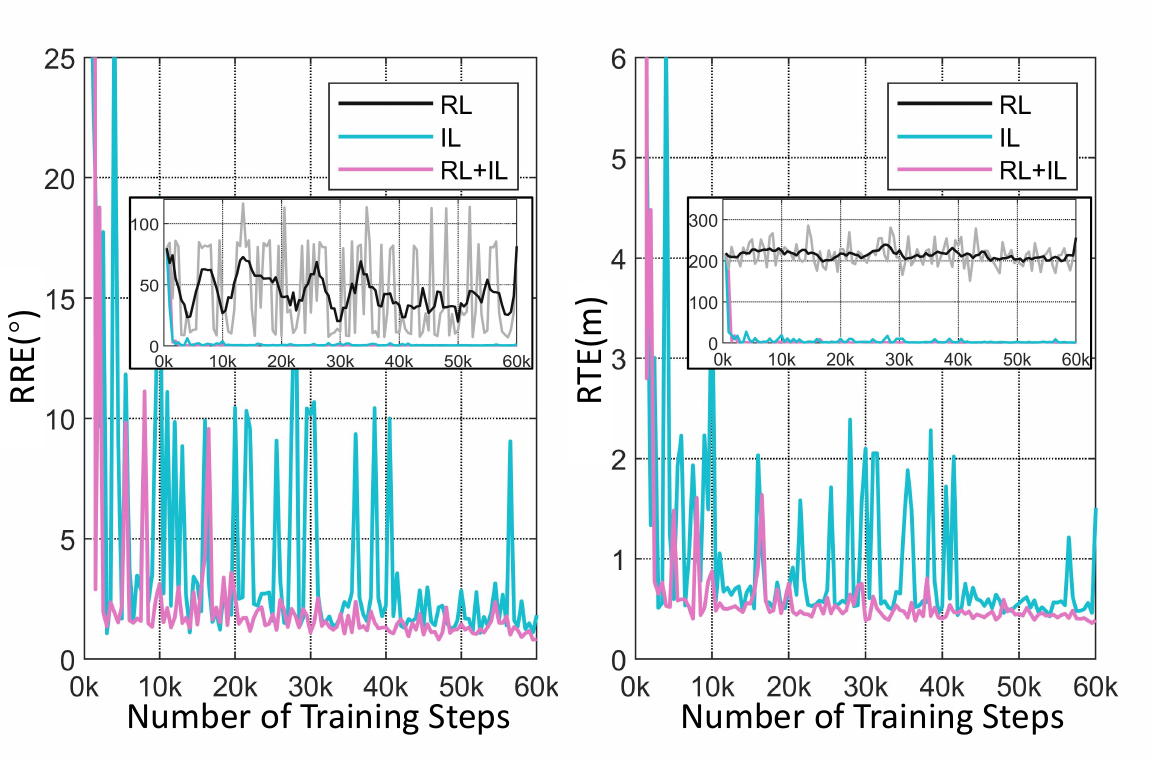}
    \vspace{-8mm}
    \caption{Convergence of CMR-Agent with different training configurations.
    }
    \label{fig7}
\end{figure}

\begin{table}[t]
\centering
\caption{Performance with different number of iterations. The results are averaged over two sequences of KITTI-Odometry.}
\vspace{-2mm}
\label{tab4}
\begin{tabular}{c|c|c|c|c}
\hline
Iteration        & RTE(m)          & RRE($^\circ$)      & RR(\%) & Time(ms) \\ \hline
1             & 2.272$\pm$1.141 & 4.346$\pm$2.800    & 23.19  & 0.035\\ 
2             & 1.213$\pm$1.008 & 3.604$\pm$2.756    & 52.83  & 0.038\\ 
3             & 0.663$\pm$0.637 & 2.649$\pm$2.479    & 78.90  & 0.042\\ 
4             & 0.426$\pm$0.427 & 1.844$\pm$2.082    & 95.47  & 0.046\\ 
5             & 0.291$\pm$0.261 & 1.138$\pm$1.404    & 99.46  & 0.049\\ 
6             & 0.236$\pm$0.193 & 0.814$\pm$1.087    & 99.70  & 0.053\\ 
7             & 0.222$\pm$0.186 & 0.691$\pm$0.976    & 99.75  & 0.057\\ 
8             & 0.218$\pm$0.187 & 0.679$\pm$0.980    & 99.77  & 0.061\\ 
9             & 0.217$\pm$0.178 & 0.658$\pm$0.956    & 99.80  & 0.064\\ 
10            & 0.217$\pm$0.181 & 0.653$\pm$0.959    & 99.80  & 0.068\\ \hline
\end{tabular}
\end{table}

\subsection{Ablation studies}
\textbf{Individual Components:} We investigate the contribution of individual components by removing them from the full pipeline of CMR-Agent. The results are listed in Table \ref{tab3}. By simply removing the 3D state representation, we can see that RR has dropped significantly compared to the other three rows. This serves as a significant evidence that neutral states (see Section. \ref{sec:state}) exist and fail to provide useful information for action decision. By simply removing the 2D state representation, we can see that RR still maintains a relatively high value, while RTE and RRE have significantly increased. This demonstrates that the 3D state supplements useful information to 2D neutral states, yet its contribution to fine registration is less than that provided by the fine-grained features of RGB images. Thus, the results in the first two rows of Table \ref{tab3} demonstrate the necessity of the 2D-3D hybrid state representation. In addition, we also investigate the role of reinforcement learning (RL) and imitation learning (IL) by training our CMR-Agent both separately and jointly. The curves of validation error are displayed in Fig .\ref{fig7}. When trained solely with RL, in line with the prior experience \cite{Shao_2020_CVPR, bauer2021reagent}, the convergence process is very slow. IL promotes quick convergence but suffers from training instability and limited accuracy. The main bottleneck lies in the greedy expert policy, and theoretically, it is impossible for the agent to learn a policy better than the expert. As shown in Fig.\ref{fig7} and the last two rows of Table. \ref{tab3}, the combination of IL and RL results in training that is both rapid and stable, breaking through the accuracy limitation of IL.

\textbf{Number of Iterations:} We iterate CMR-Agent for different numbers of times and report the results in Table \ref{tab4}. Obviously, the registration performance positively correlates with the number of iterations, and CMR-Agent only takes 3$\sim$4 ms per iteration once the one-shot embeddings are completed. A dynamic demonstration of the entire iterative process is included at the end of the supplementary video.

\section{Conclusion}
In this paper, we propose a novel cross-modal agent (CMR-Agent) for image-to-point cloud registration. Our key insight is to mimic the natural human process of aligning two objects through observation and adjustment. By reformulating the problem as a Markov decision process, we implement CMR-Agent with reinforcement learning. The proposed 2D-3D hybrid state representation fully exploits the fine-grained features of RGB images while reducing the useless neutral states. The proposed point-to-point alignment reward bridges the modality gap between images and point clouds to guide the training. The proposed framework for reusing one-shot embeddings also makes CMR-Agent very efficient. Extensive experiments demonstrate notable improvements in registration performance, showcasing the great potential of our method to achieve accurate camera localization in pre-built LiDAR maps.

\bibliographystyle{IEEEbib}
\bibliography{iros2024}

\end{document}